\documentclass[NewProceedings, a4paper, 12pt]{ascelike-new}
\usepackage{ragged2e}
\usepackage{setspace}
\usepackage{makecell}
\usepackage{float}
\usepackage[utf8]{inputenc}
\usepackage[T1]{fontenc}
\usepackage{lmodern}
\usepackage{graphicx}
\usepackage{amsmath}
\usepackage{dsfont}
\usepackage{newtxtext,newtxmath}
\usepackage[colorlinks=true,citecolor=red,linkcolor=black]{hyperref}
\usepackage[mathscr]{euscript}

\usepackage{indentfirst}
\usepackage{pgfplots}
\pgfplotsset{compat=1.14}
\usepackage{authblk}
\begin{document}
    \newgeometry{left=1.28in,right=1.28in,top=1.378in,bottom=1.378in}
\sloppy
% You will need to make the title all-caps
\title{Combining Policy Gradient and Safety-Based Control for Autonomous Driving}
\author[1*]{Xi Xiong}
\author[2]{Lu Liu}

\affil[1]{(Corresponding Author) Key Laboratory of Road and Traffic Engineering, Ministry of Education, Tongji University, Shanghai, China; E-Mail: xi{\_}xiong@tongji.edu.cn}
\affil[2]{Key Laboratory of Road and Traffic Engineering, Ministry of Education, Tongji University, Shanghai, China; E-Mail: luliu0720@tongji.edu.cn}

%\affil[2]{Third affiliation address}
\maketitle

% Please include an abstract:
\begin{abstract}
With the advancement of data-driven techniques, addressing continuous control challenges has become more efficient. However, the reliance of these methods on historical data introduces the potential for unexpected decisions in novel scenarios. To enhance performance in autonomous driving and collision avoidance, we propose a symbiotic fusion of policy gradient with safety-based control. In this study, we employ the Deep Deterministic Policy Gradient (DDPG) algorithm to enable autonomous driving in the absence of surrounding vehicles. By training the vehicle's driving policy within a stable and familiar environment, a robust and efficient learning process is achieved. Subsequently, an artificial potential field approach is utilized to formulate a collision avoidance algorithm, accounting for the presence of surrounding vehicles. Furthermore, meticulous consideration is given to path tracking methods. The amalgamation of these approaches demonstrates substantial performance across diverse scenarios, underscoring its potential for advancing autonomous driving while upholding safety standards.
\end{abstract}

\section{Introduction}

% Background and challenges
The rapid evolution of autonomous driving technology has ushered in the promise of a transformative shift in transportation, with the potential to enhance road safety, reduce congestion, and improve mobility efficiency \cite{levinson2011towards}. However, this vision is accompanied by multifaceted technical challenges that need to be addressed for autonomous vehicles to operate safely and effectively in real-world environments. Data-driven techniques have emerged as fundamental tools for training autonomous agents. These approaches enable autonomous vehicles to learn from their interactions with the environment, adapting and making decisions based on past experiences  \cite{kiran2021deep}. Nevertheless, the dependence on historical data introduces a level of unpredictability, particularly in novel or dynamic situations. Thus, the need to balance data-driven learning with the stringent safety and predictability requirements of autonomous driving has become a central challenge in the field.

% Problem and contribution
In response to these challenges, this study proposes a novel approach by symbiotically fusing policy gradient with safety-based control. Our study is motivated by the need to strike a balance between data-driven learning and safety standards, creating autonomous driving systems that are not only capable of learning from data but also excel in ensuring reliability required for real-world deployment. The integration of these two approaches has the potential to advance the field, contributing to the development of safer and more reliable autonomous vehicles.

% Literature review
Previous work illuminates the extensive body of research in the domain of decision-making for autonomous vehicles. Model-based methods establish driving behavior rules through vehicle dynamics, excelling in specific scenarios but struggling with complexity and emergencies \cite{hu2020fuzzy,Kesting2007GeneralLM,xiong2021optimizing}.
Recent advancements in deep learning and reinforcement learning have witnessed rapid growth. The amalgamation of these two methods has emerged as a powerful solution for addressing high-dimensional input spaces \cite{Zhao2020SimtoRealTI}.
Notably, the Deep Q-Network (DQN) demonstrated human-level proficiency in Atari games, but challenges arose in high-dimensional action state spaces \cite{Mnih2015HumanlevelCT,Hasselt2015DeepRL}. The Deep Deterministic Policy Gradient (DDPG) emerged as a robust alternative, excelling in continuous action spaces, achieving faster speeds, and enhanced performance in autonomous vehicle control \cite{Lillicrap2015ContinuousCW}.
%Duan et al. \cite{Duan2020DRLBasedAV} tackled Autonomous Vehicle Control (AVC) problems using DQN and DDPG algorithms in a system with 200 buses, where the results favored the DDPG algorithm, showcasing faster speed and enhanced performance. 
Liu et al. \cite{Liu2018ADR} applied the DDPG algorithm to autonomous driving decisions, combining it with supervised learning and updating the actor network in line with expert strategies.
Zhu et al. \cite{zhu2020safe} proposed a speed control method based on DDPG, surpassing human drivers and model predictive control in terms of safety, efficiency, and comfort.
However, in unfamiliar scenarios, learned policies may yield unpredictable decisions, a limitation inherent to model-free methods. To address these multifaceted challenges, we incorporate the artificial potential field method \cite{Khatib1985RealTimeOA}, widely adopted in robot path planning, to avoid collisions and enhance safety.

In this paper, we present a comprehensive combined algorithm that strategically fuses the DDPG method with safety-based control. This integration capitalizes on the unique strengths of these methodologies to formulate efficient obstacle avoidance paths for autonomous vehicles while upholding the priority of keeping the vehicle closely aligned with the central road track through the incorporation of path tracking. The deployment of the DDPG algorithm empowers autonomous learning and adaptation, thereby optimizing vehicle performance in the absence of surrounding vehicles. The integration of the artificial potential field (APF) method adds a layer of collision avoidance capability, adeptly accounting for the presence of surrounding vehicles. Simultaneously, the inclusion of path tracking technology ensures unwavering adherence to the central road track, effectively mitigating the risk of route deviations and significantly elevating overall safety standards. The conducted simulation experiments on the TORCS platform comprehensively validate the effectiveness of our proposed method, affirming its potential for advancing autonomous driving while enhancing safety measures.

%In this study, we developed a combined algorithm based on DDPG, which integrates the artificial potential field method into reinforcement learning to take advantage of both to plan more efficient obstacle avoidance paths. At the same time, we deem path tracking to be indispensable in autonomous driving strategies, and we add it to the combined algorithms to prevent the vehicle from veering off the road. Moreover, the simulation experiments further validate the effectiveness of the combined algorithm. The contribution of our proposed method can be summarised as follows:
%(1) The DDPG algorithm with a reward mechanism is used to realize the auto-learning of autonomous vehicles. Setting the speed of the vehicle along the track direction as a positive reward for the DDPG algorithm to encourage the vehicle to travel fast along the specified route and improve the performance of the vehicle in different scenarios.
%(2) The artificial potential field approach is integrated with the policy gradient technique to seamlessly blend their strengths, overcoming the limitations of the model-free approach and enhancing the system's adaptability in unfamiliar scenarios.
%(3) The path tracking technology is further integrated into the combined algorithm, which ensures the autonomous vehicle stays close to the central road track, reducing the risk of deviation from the route and improving overall safety.

% The following sections as shown as follows... (one paragraph)
The subsequent sections of this paper are structured as follows. Section~\ref{Sec: background} presents the background of policy gradient and safety-based control, providing the theoretical foundation for our work. In Section~\ref{Sec: method}, we craft the algorithmic framework, integrating policy gradient with the artificial potential field method and path tracking. Section~\ref{Sec: experiment} evaluates the performance of our combined algorithm through extensive testing on the TORCS platform. In Section~\ref{Sec: conclusion}, we conclude this study and offer insights into prospective directions for future research endeavors.

\section{RELATED WORK}
\label{Sec: background}

\subsection{Policy gradient}
We consider a Markov decision process (MDP) denoted as $\mathcal{M}=\left(\mathcal{S}, \mathscr{A}, \mathcal{T}, r, \gamma \right)$, where $\mathcal{S} \subseteq \mathbb{R}^D $ is a continuous state space for the vehicle in $\mathbb{R}^D$, $\mathscr{A}$ is the action space consisting of longitude and latitude commands, $\mathcal{T}$ is the transition probability that describes the next state distribution $s' \sim \mathcal{T}\left(\cdot | s, a \right)$ at state $s \in \mathcal{S}$ when action $a \in \mathscr{A}$ is taken, $r: \mathcal{S} \times \mathscr{A} \rightarrow \mathbb{R}$ is the reward for the vehicle, and $\gamma \in \left(0, 1\right)$ is the discount factor. We consider the vehicle speed projected along the track direction as the reward at time $t$. The goal is to find the optimal policy $\pi^*: \mathcal{S} \rightarrow \mathscr{A}$ to maximize the accumulative rewards $\mathbb{E}_{\pi^*} \left[\sum_{t=0}^{\infty} \gamma^t r\left(s_t, a_t\right)\right]$.
The state-action value function is used to represent the expected reward starting from state $s$ and taking action $a$, $Q^{\pi} \left(s, a\right) = \mathbb{E}_{\pi} \left[\sum_{t=0}^{\infty} \gamma^t r\left(s_t, a_t\right) | s_0 = s, a_0 = a\right]$. The optimal state-action value function is $Q^{*}(s,a) = \max_{\pi} Q^{\pi} \left(s, a\right)$. $Q^{*}(s,a)$ can be expressed with the Bellman equation,
\begingroup
\setlength{\abovedisplayskip}{1\baselineskip}
\setlength{\belowdisplayskip}{1\baselineskip}
\begin{align*}
    Q^*(s_t,a_t) = \mathop{\mathbb{E}} \left[r_t + \gamma \max_{a_{t+1} \in \mathbf{\mathscr{A}}}  Q^*(s_{t+1},a_{t+1}) | s_t, a_t \right].
\end{align*}
\endgroup
%The optimal action-value function can be expressed using the Bellman optimality equation.
%\begin{gather}
%    Q^*(s_t,a_t) = \mathop{\mathbb{E}} \left[r_t + \gamma \max_{a_{t+1} \in \mathbf{\mathcal{A}}}  Q^*(s_{t+1},a_{t+1}) | s_t, a_t \right].
%\end{gather}
%The objective is to find the optimal policy $\pi^\ast$ to maximize the total discounted reward.
%\subsection{Policy Gradient}

%The policy gradient algorithm can solve optimal policies in high-dimensional state spaces and high-dimensional continuous action spaces. In the DRL problem, a deep neural network with parameter $\theta$ can be used to parametrically represent the policy, and the policy gradient method can be used to optimize the policy.

A stochastic policy $\pi \left(a|s \right)$ determines the action $a \in \mathscr{A}$ for the vehicle to take given its current state $s \in \mathcal{S}$.
The process of solving a stochastic policy gradient necessitates the integration of state and action distributions, which, in practice, often entails collecting a vast num-ber of samples. In contrast, when dealing with a deterministic policy, it consistently produces a single, deterministic action in response to the same state \cite{Sutton2005ReinforcementLA}. This characteristic is particularly advantageous since the deterministic policy gradi-ent algorithm demands fewer sample data and, thus, operates more efficiently. We use the deterministic policy $a=\mu_\theta\left(s\right)$ to represent the reflection from the state spaces to the action spaces. Similar to the policy gradient theorem \cite{Sutton1999PolicyGM}, We define the objective performance as,
\begingroup
\setlength{\abovedisplayskip}{1\baselineskip}
\setlength{\belowdisplayskip}{1\baselineskip}
\begin{align*}
    J(\mu_\theta) = \int_{s}\rho^\mu(s)\cdot Q^\mu(s,a)ds=\int_{s}\rho^\mu(s)\cdot Q^\mu(s,\mu_\theta(s))ds,
\end{align*}
\endgroup
where $\rho^\mu\left(s\right)$ denotes the state distribution. If $\nabla_\theta\mu_\theta\left(s\right)$ and 
$\nabla_a Q^\mu\left(s,a\right)$ both exist, we can also derive the deterministic policy gradient theorem.
\begingroup
\setlength{\abovedisplayskip}{1\baselineskip}
\setlength{\belowdisplayskip}{1\baselineskip}
\begin{align*}
    \nabla_\theta J(\pi_\theta) = \int_{s}\rho^\mu(s)\cdot \nabla_\theta \mu_\theta(s)\cdot \nabla_a Q^\mu(s,a)ds= \mathop{\mathbb{E}} \left[\nabla_a Q^\mu(s,a)\cdot \nabla_\theta \mu_\theta(s) \right]
\end{align*}
\endgroup

\subsection{Safety-based control}

Vehicle safety is determined by various factors, with collision avoidance and maintaining a defined trajectory being paramount, particularly the former. The artificial potential field method stands out as a well-established and efficient approach in path planning, finding extensive application in research related to robot path planning  \cite{Rostami2019ObstacleAO}. This method primarily aims to discover the optimal collision-free path from the robot's initial position to its destination within an environment containing obstacles. The fundamental concept involves constructing an artificial potential field within the robot's mobile environment, influenced jointly by the attractive field at the target destination and the repulsive field generated by obstacles. The combined attractive and repulsive forces guide the robot's movements, ensuring obstacle avoidance while reaching the target. The total artificial potential field is defined as,
\begingroup
\setlength{\abovedisplayskip}{1\baselineskip}
\setlength{\belowdisplayskip}{1\baselineskip}
\begin{align*}
    U_{art}({x})=U_{att}(x)+U_{rep}(x),
\end{align*}
\endgroup
where $U_{art}({x})$ is the artificial potential field,  $U_{att}(x)$ is the attractive potential field, and $U_{rep}(x)$ is the repulsive potential field.
%\begin{gather}
%U_{art}({x})=U_{att}(x)+U_{rep}(x) \\[2ex]
%U_{att}=\frac{1}{2} k_t d^2(x,x_t) \\[2ex]
%U_{rep}= 
%    \left\{   
%            \begin{array}{lr}
%            \frac{1}{2} k_o[\frac{1}{d(x,x_o)}-\frac{1}{d_o}]^2, &d(x,x_o) \leq d_o  \\
%            0, &d(x,x_o) >  d_o 
%            \end{array}
%    \right.
%\end{gather}
%where  $\boldsymbol{U}_{art}(\boldsymbol{x})$ is the artificial potential field,  $\boldsymbol{U}_{att}(\boldsymbol{x})$ is the attractive potential field, and the $\boldsymbol{U}_{rep}(\boldsymbol{x})$ is the repulsive potential field;  $k_t$ and $k_o$ are the attractive and repulsive potential field coefficients; $\boldsymbol{x}$, $\boldsymbol{x_t}$, and $\boldsymbol{x_o}$ are the positions of agent, target, and obstacle, respectively;  $d(\boldsymbol{x},\boldsymbol{x_t})$ denotes the Euclidean distance between the agent and target,  $d(\boldsymbol{x},\boldsymbol{x_o})$ denotes the Euclidean distance between agent and obstacle,  $d_o$ represents the maximum impact distance of the obstacle.
These potential fields serve as the basis for calculating the potential forces, which are, in turn, the gradients of their respective potential fields,
\begingroup
\setlength{\abovedisplayskip}{1\baselineskip}
\setlength{\belowdisplayskip}{1\baselineskip}
\begin{align*}
    F_{att}=-\nabla U_{att}(x), \\[2ex] 
    F_{rep}= -\nabla U_{rep}(x).
\end{align*}
\endgroup

The artificial potential field method is a valuable tool for vehicle collision avoidance \cite{Huang2020AMP}. When addressing scenarios involving multiple targets and obstacles, as depicted in Figure 1, the ultimate potential force results from the vector sum of attractive and repulsive forces. Specifically, the attractive forces, denoted as $F_{att1}$ and $F_{att2}$, originate from Target 1 and Target 2, with $F_{att}$ being the combined vector sum of $F_{att1}$ and $F_{att2}$. Simultaneously, the repulsive force $F_{rep}$ is the amalgamation of $F_{rep1}$ and $F_{rep2}$, which are generated by Obstacle 1 and Obstacle 2, respectively. This approach allows for comprehensive collision avoidance and navigation in the presence of multiple targets and obstacles.

\begin{figure}[H]
    \centering
    \includegraphics[width=0.8\linewidth,trim=100 60 100 40,clip]{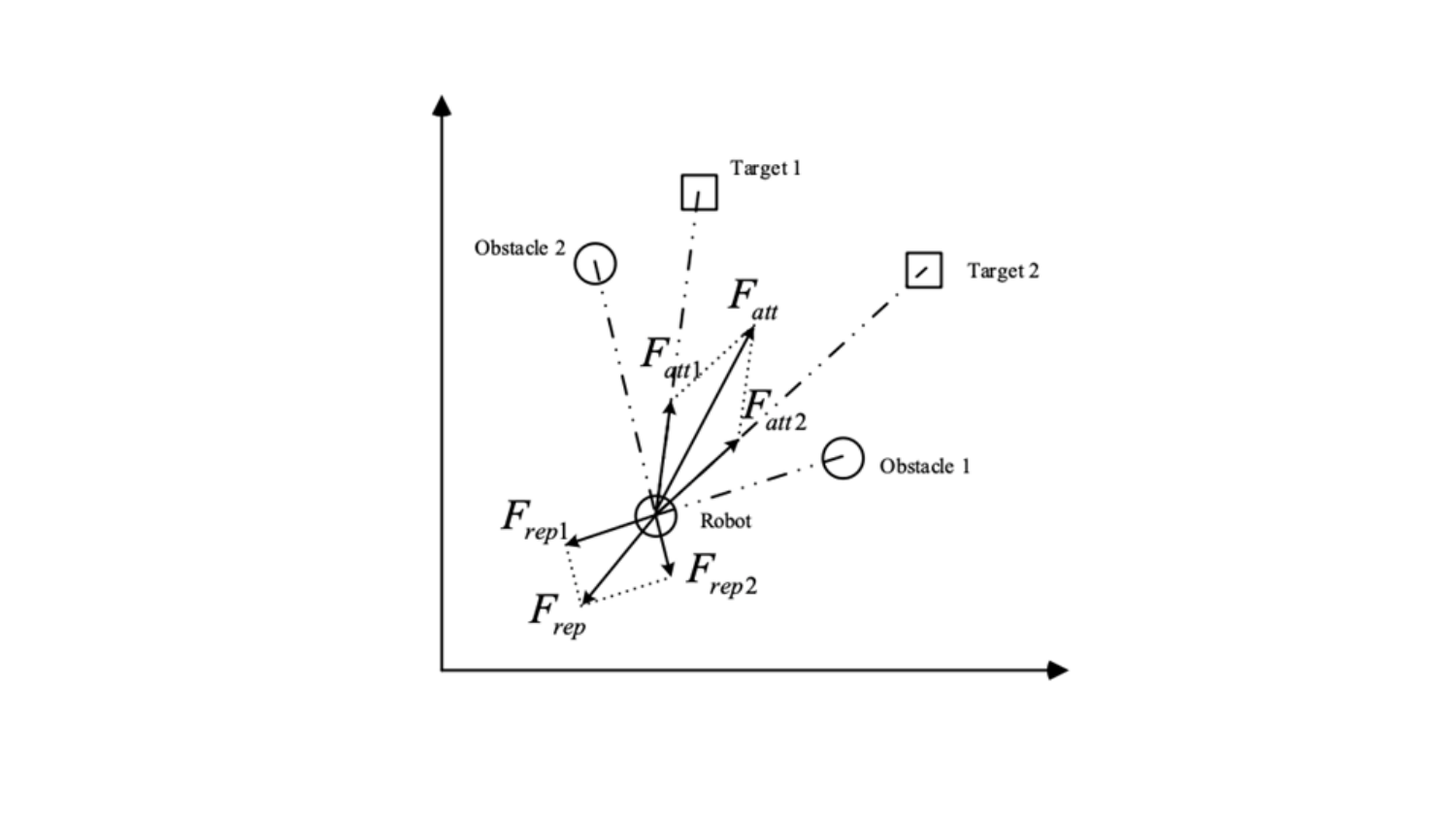}
    \caption{The artificial potential field with multiple targets and obstacles1.}
    \label{Figure 1.}
\end{figure}

\section{Methodology}
\label{Sec: method}

\subsection{Framework}

We make use of perception sensor data, including details like vehicle speed, position on the road track, and distances to opponent vehicles \cite{Loiacono2013SimulatedCR}. The input data is categorized into two distinct components, with opponent distances influencing collision avoidance and other parameters guiding policy gradient and path tracking. Notably, the system outputs steering and acceleration actions. The initial training phase ensures the vehicle's efficient operation in a stable environment devoid of opponent vehicles. Simultaneously, we introduce the artificial potential field method, which adjusts the vehicle's steering and acceleration responses based on repulsion potential fields when obstacles are present. The overarching algorithm integrates the path tracking method to maintain the vehicle's trajectory along the road's central track.

Our algorithm comprises three key methods: the deterministic policy gradient method, the artificial potential field method, and the path tracking method. Each of these methods yields its own set of acceleration and steering commands. To obtain the vehicle's final steering and acceleration actions, we employ a weighted combination of these three action outputs, expressed as follows:
\begingroup
\setlength{\abovedisplayskip}{1\baselineskip}
\setlength{\belowdisplayskip}{1\baselineskip}
\begin{align*}
\delta=& \alpha \cdot \delta_l + \beta \cdot \delta_f + \lambda \cdot \delta_p \\[2ex]
\tau=& \alpha \cdot \tau_l + \beta \cdot \tau_f + \lambda \cdot \tau_p \\[2ex]
& s.t. \ \alpha+\beta+\lambda=1
\end{align*}
\endgroup
where $\delta$ represents the final steering action, $\delta_l$ stands for the steering action derived from policy gradient, $\delta_f$ corresponds to the steering action originating from the artificial potential field approach, and $\delta_p$ signifies the steering action involved in path tracking. Similarly, $\tau$ denotes the final acceleration action, $\tau_l$ is associated with the acceleration generated by policy gradient, $\tau_f$ is linked to the acceleration derived from the artificial potential field method, and $\tau_p$ represents the acceleration relevant to path tracking. Moreover, we introduce $\alpha$, $\beta$, and $\lambda$ as weight parameters associated with these three methods. This balanced combination of actions is designed to enable the vehicle to navigate effectively, taking into consideration the diverse requirements of autonomous driving, collision avoidance, and path tracking.

\subsection{Deterministic policy gradient}
We employ the deterministic policy gradient for autonomous driving control in the absence of opponents. At time $t$, the state $s_t$ comprises various parameters, including vehicle speed, engine speed, track position, and vehicle angle. The action $a_t$ encompasses both the steering action $\delta_l^t$ and acceleration action $\tau_l^t$. The reward function, denoted as $r(s_t, a_t)$, quantifies the vehicle's speed projected along the track direction. To implement the deterministic policy gradient algorithm, we draw inspiration from the actor-critic framework presented in \cite{Lillicrap2015ContinuousCW}. In this framework, the actor network generates the action $a_t$ based on the state $s_t$ at time $t$, which is then used to interact with the environment to produce $s_{t+1}$ and calculate the reward $r(s_t, a_t)$. The actor network selects the optimal action, while the critic network evaluates the quality of actions. The critic network updates its parameters $\omega^Q$ by minimizing the loss function,
\begingroup
\setlength{\abovedisplayskip}{1\baselineskip}
\setlength{\belowdisplayskip}{1\baselineskip}
\begin{align}
    \label{Eq: value function}
    L(\omega^Q) = \mathop{\mathbb{E}} \left[(r(s_t,a_t) + \gamma Q'(s_{t+1},\mu'(s_{t+1});\omega^{Q'})-Q(s_t,a_t;\omega^Q))^2  \right].
\end{align}
\endgroup
Furthermore, the actor computes the gradient of $Q(s_t, a_t)$ and updates the parameters $\omega^{\mu}$ in the direction that maximizes the Q-value,
\begingroup
\setlength{\abovedisplayskip}{1\baselineskip}
\setlength{\belowdisplayskip}{1\baselineskip}
\begin{align}
    \label{Eq: policy function}
    \nabla_{\omega^\mu} J=\mathop{\mathbb{E}} \left[\nabla_a Q^\mu(s,a)\cdot \nabla_{\omega^\mu} \mu(s)  \right].
\end{align}
\endgroup
Deep neural networks are employed to approximate both the value function $Q(s,a)$ and the policy $\mu(s)$. The parameters $\omega^Q$ are iteratively updated by computing the gradient of the loss function $\nabla_{\omega^Q} {L(\omega^Q)}$ as defined in Eq.~\ref{Eq: value function}. Similarly, the parameters $\omega^\mu$ are updated by determining the gradient of the objective function $\nabla_{\omega^\mu} J$ as described in Eq.~\ref{Eq: policy function}.

\subsection{Artificial potential field}

With respect to the artificial potential field, our primary emphasis is directed towards the repulsive potential field, a pivotal component in our approach that empowers autonomous vehicles to adeptly avert collisions when confronted with neigh-boring opponent vehicles in their proximity.

As depicted in Figure 2, within the ego vehicle's coordinate system, real-time position and distance data of obstacles are accessible. These data serve as the basis for computing the repulsive force, denoted as $F_{rep}$, which acts upon the vehicle. Notably, the component of $F_{rep}$ projected along the x-axis corresponds to the steering command, while the component projected along the y-axis corresponds to the acceleration command. It is important to note that we assume these forces to be continuous and solely dependent on the distances between the vehicle and obstacles,
\begin{figure}[H]
    \centering
    \includegraphics[width=0.75\linewidth,trim=120 100 100 10,clip]{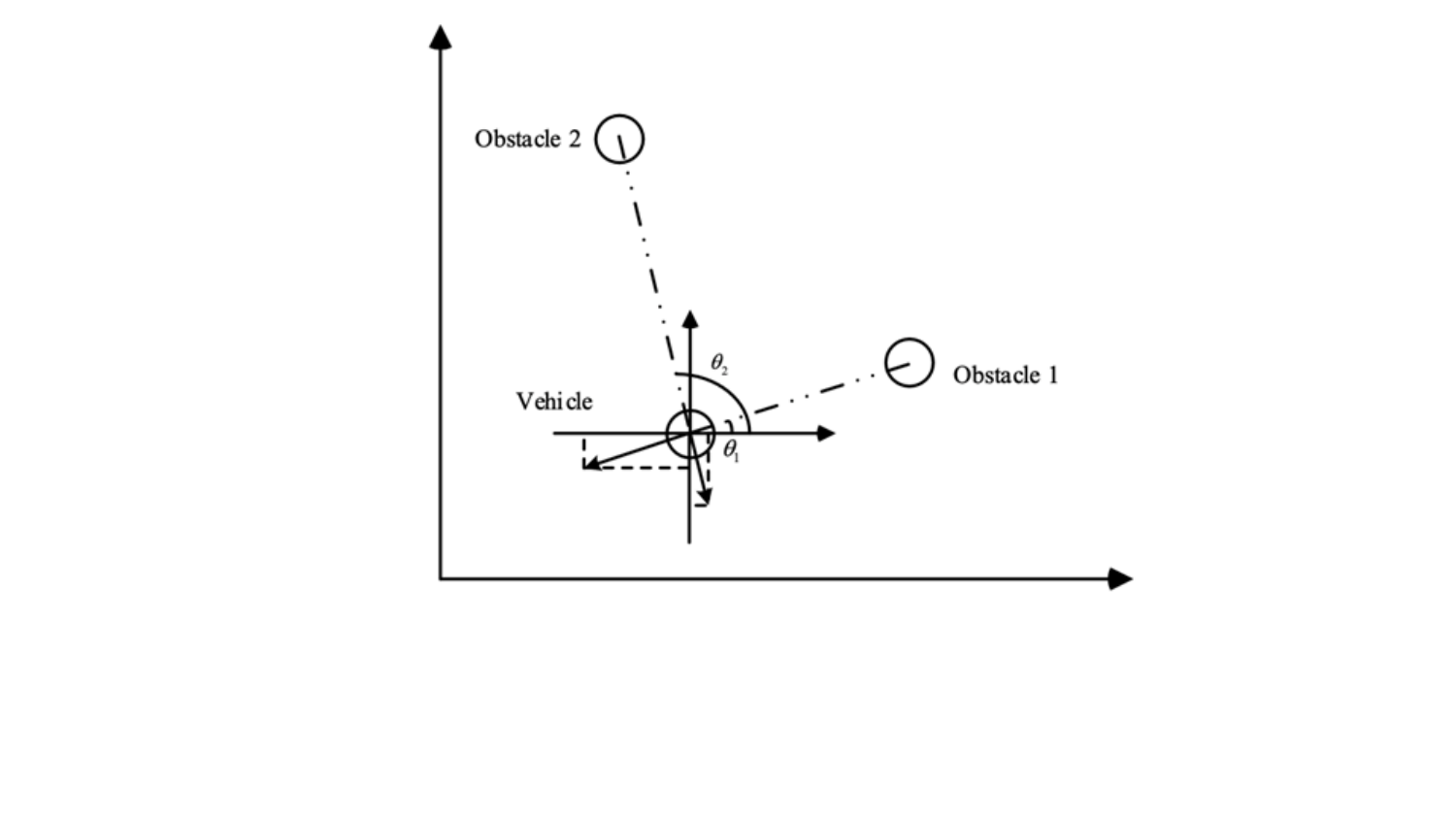}
    \caption{Repulsive potential field forces for the ego-vehicle.}
    \label{Figure 2.}
    \vspace {-2.5ex}
\end{figure}
\begingroup
\setlength{\abovedisplayskip}{-0.5\baselineskip}
\setlength{\belowdisplayskip}{1\baselineskip}
\begin{align*}
F_{{rep}_x}=-\sum_i \frac{1}{d_i^\eta} \cos\theta_i \\[2ex]
F_{{rep}_y}=-\sum_i \frac{1}{d_i^\eta} \sin\theta_i
\end{align*}
\endgroup
where $\theta_i$ signifies the angle of obstacle $i$ within the ego vehicle's coordinate system, while $d_i$ represents the distance between the ego vehicle and obstacle $i$. Additionally, the parameter $\eta$ is introduced as the exponent, allowing for its determination to influence the nature of the force calculation.
The output actions are proportional to the repulsive force,
\begingroup
\setlength{\abovedisplayskip}{1\baselineskip}
\setlength{\belowdisplayskip}{1\baselineskip}
\begin{align*}
\delta_f=k_{fx}F_{rep_x}, \\[2ex]
\tau_f=k_{fy}F_{rep_y},
\end{align*}
\endgroup
where $k_{fx}$ and $k_{fy}$ are associated with $F_{rep_x}$ and $F_{rep_y}$, respectively. These coefficients determine the scaling factors for the forces in the x and y directions.

\subsection{Path tracking}

In the context of path tracking, the primary objective is to guide the vehicle along the central track of the road. This entails minimizing the heading error between the vehicle direction and the direction of the track axis, as well as reducing the tracking error between the vehicle centroid and the central road track, as described in \cite{Kapania2015PathTO}. The heading error and tracing error are shown in Figure 3.
\begin{figure}[H]
    \centering
    \includegraphics[width=0.5\linewidth,trim=100 80 100 85,clip]{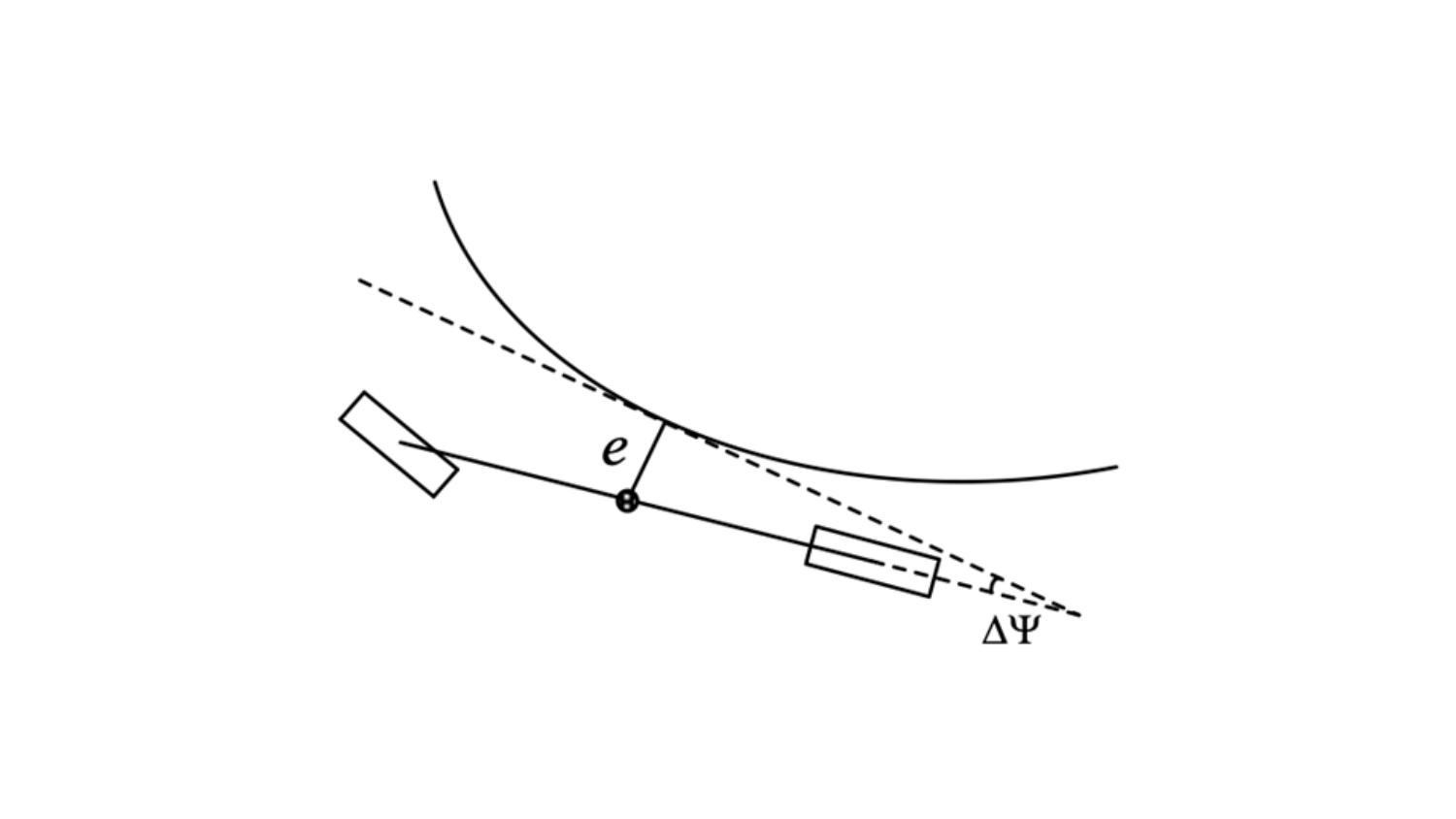}
    \caption{The heading error $\Delta \Psi$ and the tracking error $e$.}
    \label{Figure 3.}
    %\vspace {-2.5ex}
\end{figure}
The steering command, denoted as $\delta_p$, is determined based on tracking error and heading error and can be expressed as,
\begingroup
\setlength{\abovedisplayskip}{1\baselineskip}
\setlength{\belowdisplayskip}{1\baselineskip}
\begin{align*}
\delta_p = \eta_1 \cdot \Delta \Psi + \eta_2 \cdot e,
\end{align*}
\endgroup
where $\Delta \Psi$ represents the heading error, $e$ is the tracking error, and $\eta_1$ and $\eta_2$ are the respective coefficients for these terms.
The acceleration command $\tau_p$ is influenced by the steering command. In general, the rule is to reduce the vehicle's speed when the steering command reaches a sufficiently high value.
\section{Numerical Results}
\label{Sec: experiment}

\subsection{Setting}

Our approach to autonomous driving is implemented on the TORCS platform \cite{Loiacono2013SimulatedCR}. Initially, we employ the deep deterministic policy gradient (DDPG) for decision-making in scenarios where the vehicle operates without the presence of surrounding vehicles.  The input states used in the actor-critic framework include critical parameters such as vehicle speed, engine speed, wheel speed, track position, and vehicle angle. These input data guide the generation of output actions, which consist of steering and acceleration commands. Notably, the actor and critic neural networks both comprise two hidden layers, each featuring 400 and 300 units. These networks undergo training with specified learning rates of $10^{-4}$ for the actor and $10^{-3}$ for the critic, respectively. The final output layer of the actor network employs a $\tanh$ activation function to ensure precise scaling of the actions for steering and acceleration.
%train the vehicle without opponents on GPU. The input states for the actor-critic architecture are focus sensors, track sensors, vehicle speed, engine speed, wheel speed, track position, and vehicle angle. The output actions are the steering commands and acceleration commands. We set the parameters of the DDPG algorithm as follows: the discounted factor $\gamma$ is 0.99, the training minibatch size is 64; the actor and critic neural networks both have two hidden layers of size (400, 300), and the learning rates of the two networks are $10^{-4}$ and $10^{-3}$ respectively. The final output layer of the actor neural network is a tanh layer to implement steering and acceleration commands.

Subsequently, our approach seamlessly combines the actions derived from the DDPG with those originating from safety-based control strategies. The parameters guiding our approach, detailed in Table \ref{table 1}, are meticulously organized to optimize performance. In the context of the repulsive potential field, input states are derived from the distances to 36 surrounding vehicles. The path tracking method, on the other hand, utilizes angle error and track position data to precisely determine the actions, enhancing the system's ability to navigate challenging real-world scenarios effectively and safely.
\\

%36 opponent distances. The path tracking method uses the angle error and track position to determine the actions.
\begin{table}[H]
    \vspace{-0.8em}
    \centering
    \setlength{\belowcaptionskip}{0.2cm}
    \caption{\textbf{Parameters for Deterministic Policy Gradient and Safety-Based Control}}
    \label{table 1}
    \setlength{\tabcolsep}{10mm}{
    \begin{tabular}{c|c|c|c}
    \Xhline{1.2pt}
    \textbf{Parameter} & \textbf{Value} & \textbf{Parameter} & \textbf{Value} \\ 
    \hline
    $k_{fx}$        & 20             & $\eta_2$        & 2              \\
    $k_{fy}$        & 10             & $\alpha$        & 0.4            \\
    $\eta$          & 1.5            & $\beta$         & 0.3            \\
    $\eta_1$        & 3.18           & $\lambda$       & 0.3             \\ 
    \Xhline{1.2pt}
    \end{tabular}}
\end{table}

\subsection{Results and analysis}

Concerning the deterministic policy gradient for decision-making, we normalize the projected speed along the track, constraining it to the range $\left[0,2\right]$. This normalization effectively limits the one-step reward, ensuring stable and controlled learning. Figure 4 illustrates the progression of the average Q-value over the course of training. Notably, the Q-value of the critic gradually increases with time, reflecting the system's improving understanding of the environment. After roughly 13 hours of training, the average Q-value converges to approximately 110, indicating significant learning progress. Subsequently, we deploy the trained policy for autonomous driving in scenarios where no other vehicles are present. In this context, the vehicle exhibits commendable performance, demonstrating the effectiveness of the trained policy network.

\begin{figure}[H]
    \centering
    \includegraphics[width=0.8\linewidth,trim=100 70 100 70,clip]{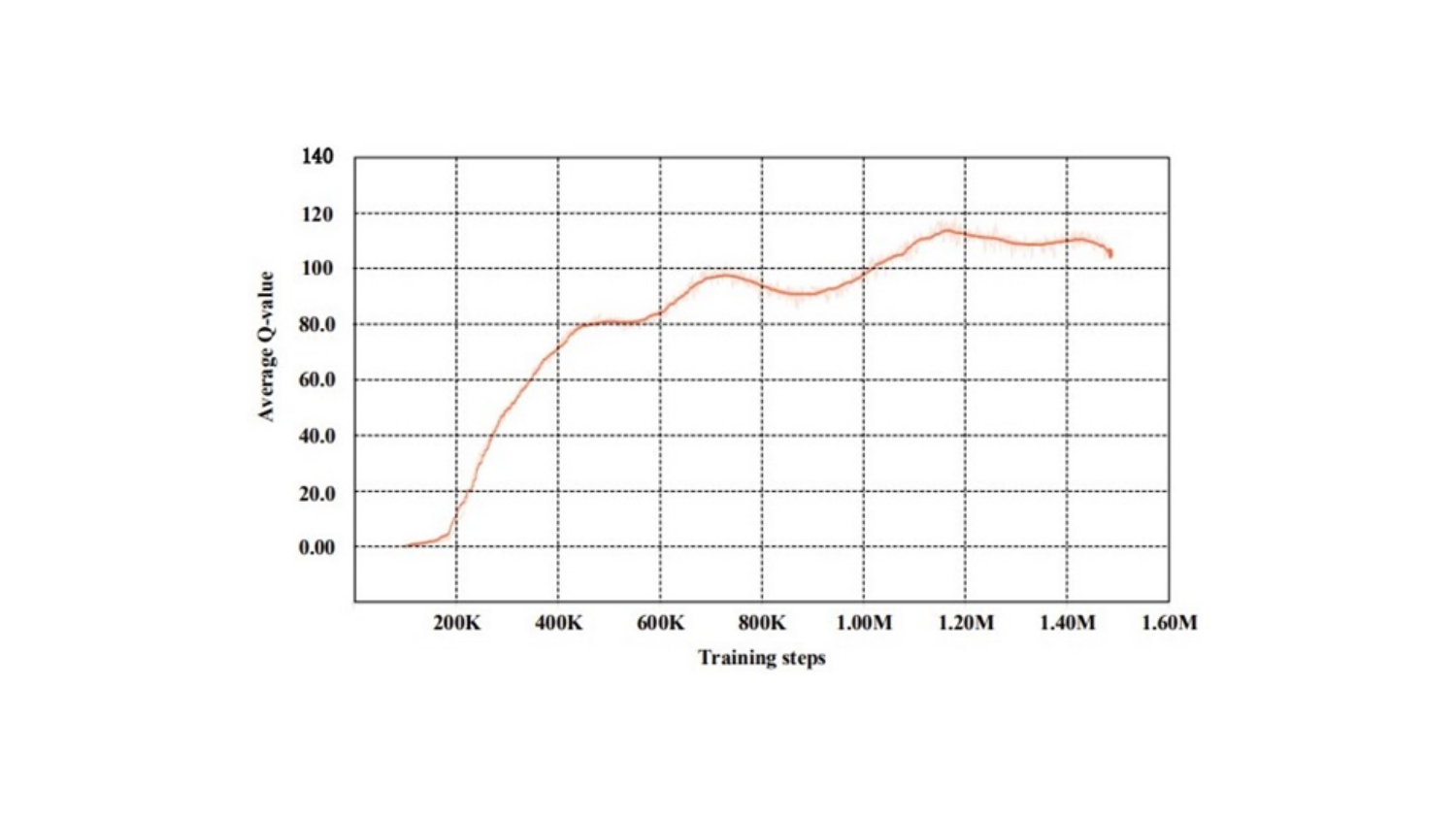}
    \caption{Average Q-value with the training steps.}
    \label{Figure 4.}
    %\vspace {-2.5ex}
\end{figure}

In Figure 5(a), we observe the vehicle smoothly navigating along a curved path. Since no surrounding vehicles are detected, the DDPG algorithm takes the lead in steering command generation. In this scenario, both the artificial potential field (APF) commands for steering and acceleration are set to 0, as there are no obstacles requiring avoidance. While the vehicle follows the curve, it deviates slightly from the central track of the road. To address this deviation, the path tracking component issues a secondary steering command. Notably, given the absence of nearby opponents and the curvature of the road, the acceleration commands for the vehicle remain close to 0. This coordinated operation demonstrates the system's effective decision-making in a dynamic environment without surrounding vehicles.

In Figures 5(b) and 5(c), we observe scenarios featuring a single nearby opponent vehicle, prompting the utilization of artificial potential field (APF) commands for safe navigation. In Figure 5(b), where the distance between our vehicle and the opponent is relatively large, the APF commands are characterized by their mild influence, resulting in subtle steering and acceleration adjustments. Concurrently, the path tracking component predominantly contributes steering commands to keep the vehicle on the central road track, while the DDPG algorithm takes the lead in managing acceleration. Conversely, in Figure 5(c), where the opponent vehicle is closer, the APF commands play a more dominant role, ensuring that the vehicle effectively avoids the nearby obstacle. These scenarios exemplify the system's dynamic adaptability, showcasing its effective collision avoidance and path-following capabilities in varying conditions.

In Figure 5(d), the autonomous vehicle is navigating along a curve while being surrounded by two other vehicles. Notably, the autonomous vehicle has deviated considerably from the central track of the road, prompting a significant steering command from the path tracking component to reorient it. Furthermore, the two opponent vehicles are positioned at a considerable distance from the autonomous vehicle, resulting in the DDPG algorithm having a more prominent role compared to the artificial potential field (APF).

The simulation experiments presented above highlight the system's remarkable adaptability. Depending on the specific scenario in which the autonomous vehicle operates, the combined algorithm adeptly manages the interplay between the three methods. This dynamic coordination is influenced by factors such as the distance to obstacles, the number of obstacles in the vicinity, and the vehicle's position. The system's ability to optimize the policy for the autonomous vehicle, ensuring safe and efficient navigation in a variety of real-world situations, underscores its versatility and robustness in responding to dynamic environmental conditions.
\begin{figure}[H]
    \centering
    \includegraphics[width=1\linewidth,trim=10 20 10 20,clip]{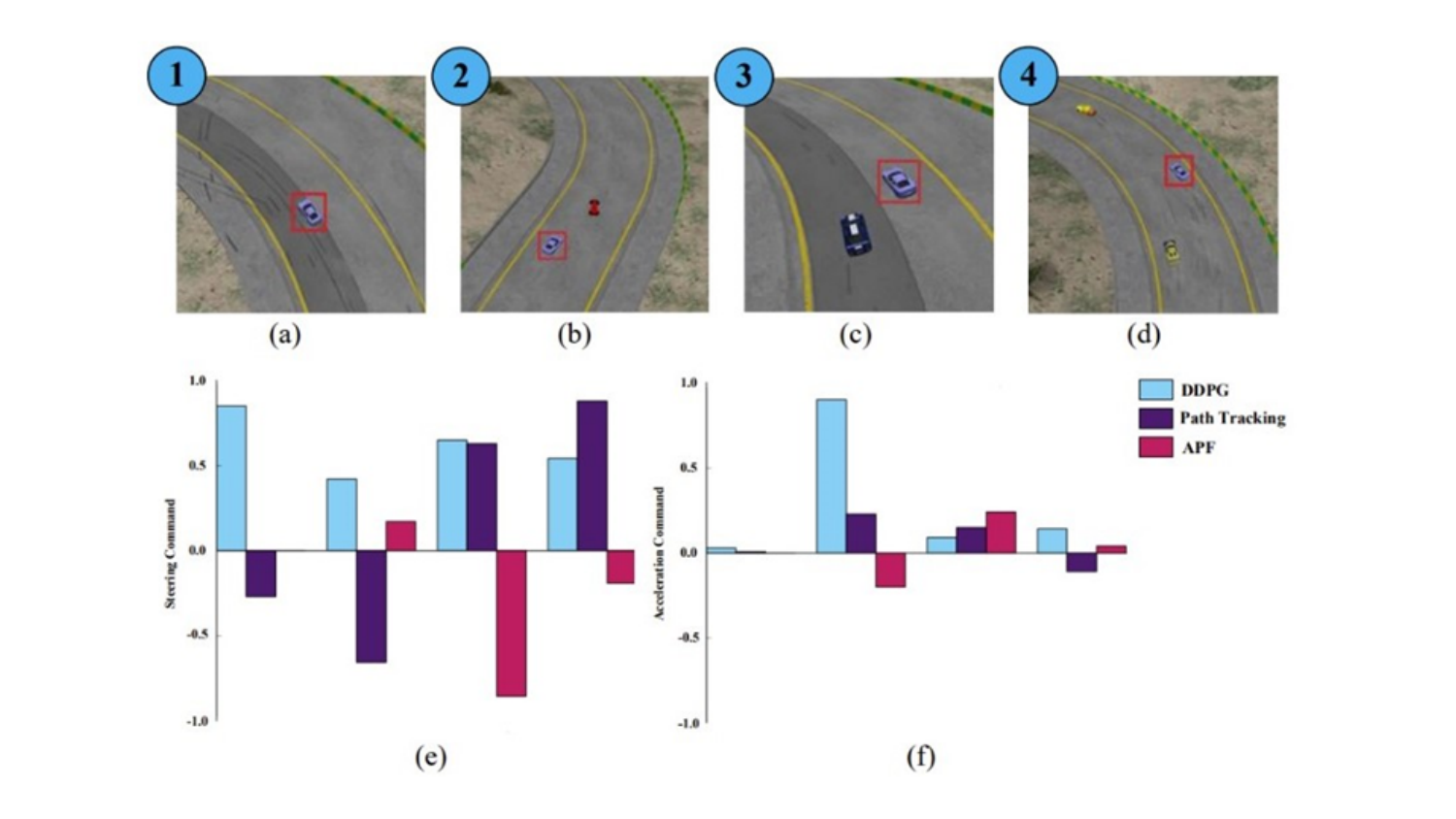}
    \caption{Typical driving scenarios and corresponding commands.}
    \label{Figure 5.}
    %\vspace {-2.5ex}
\end{figure}

\section{Conclusion}
\label{Sec: conclusion}

In this study, we have introduced a novel approach by merging the deterministic policy gradient algorithm with safety-based control mechanisms, specifically the artificial potential field method and the path tracking method. The DDPG algorithm is employed to derive a policy network for the decision-making without surrounding vehicles. This trained policy network is then seamlessly integrated with safety-based control strategies, empowering the vehicle to effectively avoid collisions and remain on track in the presence of obstacles. Our approach was rigorously evaluated through experiments conducted on the TORCS platform. The results of these experiments demonstrate the superior performance of our combined algorithm in a variety of real-world scenarios, underlining its capacity to significantly enhance the safety control measures for autonomous vehicles.

This work can be extended in several directions. First, enhancing the deterministic policy gradient with integrated vehicle dynamics can bolster data efficiency and provide more accurate modeling of vehicle behavior. Second, the exploration of multi-agent reinforcement learning techniques holds great promise for understanding and enabling effective cooperation and coordination among multiple autonomous vehicles. Third, the adoption of transformer architecture offers the potential to revolutionize the policy and value function approximation, allowing for the capture of intricate long-range dependencies in dynamic environments.

\section{ACKNOWLEDGEMENTS}
This work was supported in part by NSFC Project 72371172 and Fundamental Research Funds for the Central Universities.

%
% Here's the list of references:
%
% \label{section:references}
%\bibliography{ascexmpl-new}
%
\bibliography{ascexmpl-new}
\end{document}